\documentclass[letterpaper, 10pt]{article}
\usepackage{graphicx}
\usepackage{amsmath}
\usepackage{float}
\usepackage{amsfonts}
\usepackage[inner=0.75in, outer=0.75in, top=0.75in, bottom=1in, paperheight=11in, paperwidth=8.5in]{geometry}
\usepackage{algorithm}
\usepackage{algorithmicx}
\usepackage{algpseudocode}
\usepackage{epstopdf}
\usepackage{bm}
\usepackage{bbm}
\usepackage{tikz}
\usepackage{multirow}
\usepackage{amsmath}
\usetikzlibrary{bayesnet}
\usetikzlibrary{shapes.gates.logic.US,trees,positioning,arrows}
\usepackage{caption}
\usepackage{subcaption}
\usetikzlibrary{shapes,snakes}
\usetikzlibrary{trees}
\usepackage{cite}

\usepackage{amsthm}

\usepackage{epstopdf}
\usepackage{amssymb}
\usepackage{microtype}
\usepackage{url}
\usepackage{caption}
\usepackage{subcaption}

\usepackage[hidelinks]{hyperref}
\usepackage[capitalize]{cleveref}

\usepackage{xparse}

\usepackage[acronym,toc,nonumberlist]{glossaries}

\newacronym{SHM}{SHM}{Structural Health Monitoring}
\newacronym{PBSHM}{PBSHM}{Population-Based Structural Health Monitoring}
\newacronym{knn}{$k$-NN}{$k$-Nearest Neighbour}
\newacronym{tca}{TCA}{Transfer Component Analysis}
\newacronym{jda}{JDA}{Joint Domain Adaption}
\newacronym{artl}{ARTL}{Adaptation Regularization based Transfer Learning}
\newacronym{RKHS}{RKHS}{Reproducing Kernel Hilbert Space}
\newacronym{mmd}{MMD}{Maximum Mean Discrepancy}
\newacronym{pca}{PCA}{Principle Component Analysis}
\newacronym{SVM}{SVM}{Support Vector Machine}

\newcommand{\trace}[1]{\text{tr}{\left(#1\right)}} % trace

\DeclareDocumentCommand{\pof}{m g}{ % Probability of 
	{p( #1 %
		\IfNoValueF{#2}{\,\vert\, #2}%
		)%
	}
}

\DeclareDocumentCommand{\expect}{m g}{ % Expectation
	{\mathbb{E}\IfNoValueF{#2}{_{#2}}\left( #1 \right)%
	}
}

\DeclareDocumentCommand{\gaussianDist}{m m g}{ % Normally distributed 
	{\mathcal{N}\left(\IfNoValueF{#3}{#3 \, \vert \,}
		#1,#2\right)%
	}
}
\begin{document}

	\title{\bf Quantifying the value of information transfer in population-based SHM}
	\author{A.J.\ Hughes$^1$, J.\ Poole$^{1,2}$, N.\ Dervilis$^1$, P.\ Gardner$^3$ \& K.\ Worden$^{1,2}$ \\ ~ \\
    $^1$Dynamics Research Group, Department of Mechanical Engineering, University of Sheffield, \\ Mappin Street, Sheffield S1 3JD, UK \\
    $^2$The Alan Turing Institute, The British Library, 96 Euston Road, London, NW1 2DB, UK \\
    $^3$Frazer-Nash Consultancy, Warrington, UK
	}
	\date{}
	\maketitle
	\thispagestyle{empty}
	
	\section*{Abstract}
	
	Population-based structural health monitoring (PBSHM), seeks to address some of the limitations associated with data scarcity that arise in traditional SHM. A tenet of the population-based approach to SHM is that information can be shared between sufficiently-similar structures in order to improve predictive models. Transfer learning techniques, such as domain adaptation, have been shown to be a highly-useful technology for sharing information between structures when developing statistical classifiers for PBSHM. Nonetheless, transfer-learning techniques are not without their pitfalls. In some circumstances, for example if the data distributions associated with the structures within a population are dissimilar, applying transfer-learning methods can be detrimental to classification performance – this phenomenon is known as \textit{negative transfer}. When considered in the context of operation and maintenance decision processes, negative transfer has significant implications. Deterioration in classification performance could translate to unnecessary inspections or repairs, and even critical maintenance interventions being missed entirely. Such changes in operation and maintenance strategy would result in additional costs being incurred and could undermine the integrity and safety of structures. Given the potentially-severe consequences of negative transfer, it is prudent for engineers to ask the question “when, what, and how should one transfer between structures?”.
	
	The current paper aims to demonstrate a transfer-strategy decision process for a classification task for a population of simulated structures in the context of a representative SHM maintenance problem, supported by domain adaptation. The transfer decision framework is based upon the concept of expected value of information transfer. In order to compute the \textit{expected value of information transfer}, predictions must be made regarding the classification (and decision performance) in the target domain following information transfer. In order to forecast the outcome of transfers, a probabilistic regression is used here to predict classification performance from a proxy for structural similarity based on the modal assurance criterion.

\textbf{Keywords: value of information; transfer learning; risk; population-based SHM}
	
	\section{Introduction}
	
	Structural health monitoring (SHM) systems are technologies that acquire, process, and analyse data in order to detect, characterise, and forecast damage within structures and infrastructure across a variety of engineering domains including mechanical, civil, and aerospace\cite{Farrar2013, Rytter1993}. Fundamentally, SHM systems can be considered as decision-support tools; the information they provide regarding the current and future health states can be used to optimise the design, operation, and maintenance of structures resulting in improved safety, financials, and the extension of operational life past design specification.

	Recently, structural health monitoring systems have been viewed explicitly in the context of operation and maintenance (O\&M) decision-making \cite{Papakonstantinou2014a,Schobi2016,Vega2020a,Hughes2021,Arcieri2023bridging}. Within these works, several submodels have been identified as elements necessary to adequately define O\&M decision processes for structures. These submodels include observation models for inferring current estimates of health-states and transition models that forecast future health-states under various courses of action. Notably, these submodels require labelled data for learning and/or validation such that engineers can be confident in the information provided and the decisions recommended by SHM systems.

	One of the critical challenges associated with the development of SHM systems is the scarcity of the data necessary for the learning and validation of models. Prior to the implementation of a monitoring system, there is often a lack of comprehensive labelled data across the health-states of interest for a given structure, as obtaining data corresponding to damage states tends to  be prohibitively expensive or otherwise infeasible. Population-based structural health monitoring (PBSHM), provides a holistic framework for overcoming data scarcity in the development of predictive models for SHM \cite{Bull2021,Gosliga2021,Gardner2021b,Tsialiamanis2021}. The principal tenet of the population-based approach to SHM is that information can be shared, or transferred, between sufficiently-similar structures in order to improve predictive models. Moreover, it is posited that the more similar two structures are, the more beneficial information sharing will be between the two structures in terms of improving predictive capabilities -- preliminary research suggests that this thesis holds true \cite{Poole2023negative}. To this end, a core area of research within PBSHM is in finding reduced-order representations, and corresponding distances measures for such representations, that facilitate the quantification of structural similarity \cite{Gosliga2021,Brennan2023comparison}.

	Several transfer learning approaches have been demonstrated to be successful in sharing information between structures in the context of PBSHM, including but not limited to, a suite of algorithms that can be categorised as \textit{domain adaptation} methods \cite{Poole2022alignment,Gardner2022aircraft}. Although transfer-learning techniques have seen some success, they are not infallible. In particular, transfer learning approaches are susceptible to a phenomenon known as \textit{negative transfer} \cite{Wang2019negative}. Negative transfer is when the application of transfer learning results in a deterioration of prediction performance for a given model. When considered in the context of the operation and maintenance decision-making associated with (population-based) SHM, negative transfer has serious implications; at best it results in otherwise avoidable expenditure in the form of unnecessary inspections and/or repairs, and at worst it may result in missed inspections and repairs, undermining structural safety in a manner that is hazardous to human life.
	
	The work presented in the current paper seeks to bridge the research gap between the assessment of structural similarity and transfer learning for PBSHM, with particular a focus on decision-making. This aim is accomplished by using a numerical case study to demonstrate how the correlations between structural similarity and post-transfer prediction capabilities can be exploited to minimise the risk of negative transfer. This minimisation is achieved by extending a notion that arises in decision theory known as \textit{value of information} to transfer learning; the resulting quantity is termed the \textit{value of information transfer}. It is shown that, by optimising a transfer strategy with respect to the expected value of information transfer, one can minimise the risk of negative transfer.

	The layout of the current paper is as follows. Brief introductions to PBSHM, and transfer learning, and risk-informed transfer are provided in Section \ref{sec:background}. Following these introductions, a case-study population comprising varying 10 degree-of-freedom systems is established in Section \ref{sec:case_study}. Subsequently, the results arising from the case study are presented in Section \ref{sec:results}. Points of interest and limitations of the approach are highlighted in Section \ref{sec:Discussion}. Finally, conclusions are provided in Section \ref{sec:conclusions}.

	\section{Background}\label{sec:background}
	
	\subsection{Population-based SHM}
	
	The foundations of PBSHM have been presented in a series of journal papers, each detailing the fundamental concepts of the approach; homogeneous populations \cite{Bull2021}, heterogeneous populations \cite{Gosliga2021}, mapping and transfer \cite{Gardner2021b}, and the geometric spaces in which structures exist \cite{Tsialiamanis2021} -- for a comprehensive explanation of the ideas underpinning PBSHM, the reader is directed to these papers.
	
	As mentioned previously, adopting a population-based approach to SHM yields the potential for improved diagnostic and prognostic capabilities \cite{Worden2020}. Within a population, structures may share common characteristics such as geometries, topologies, materials, and boundary conditions; thus, it is hypothesised that as the structural similarity between individuals in a population increases, the more useful the information from one structure will be for making inferences about the other. One of the key active research areas in PBSHM is in representing structures in a manner that facilitates the quantification of structural similarity. In particular, graph-based methods have been shown to be highly effective at this \cite{Gosliga2021}, having previously been used successfully in fields such as chemistry and proteomics.
	
	When discussing information-sharing in PBSHM, it is useful to consider two categories of populations: \textit{homogeneous} populations, and \textit{heterogeneous} populations. Variation between structures in homogeneous populations may arise because of factors such as environmental conditions and manufacturing defects. In essence, \textit{heterogeneous} populations form the complement of the set of homogeneous populations \cite{Worden2020}; that is, heterogeneous populations are not exlusively comprised of sturcures that are nominally identical. Heterogeneous populations represent more general sets of structures and allow for differing designs, large variability in boundary conditions, and even multiple types of structure. While there may be stark differences between individual structures in a heterogeneous population, there may nonetheless be similarities that can be exploited to achieve useful knowledge and information transfer.

	\subsection{Transfer Learning for PBSHM}\label{sec:transfer}

	As stated previously, the key benefit in taking a population-based approach to SHM is gaining the ability to share information between sufficiently-similar structures; thereby overcoming issues with data scarcity typically associated with traditional SHM. One of the approaches for information sharing that has seen significant success is known as \textit{transfer learning}.

	Transfer-learning techniques aim use information from a data-, or label-, rich \textit{source domain} $\mathcal{D}_s = \{ \mathbf{x}_{s,i}, y_{s,i} \}_{i=1}^{N}$, in order to improve predictions in a data-, or label-, scarce \textit{target domain} $\mathcal{D}_t = \{\mathbf{x}_{t,j},y_{t,j}\}_{j = 1}^{M}$. In the context of PBSHM, it is assumed that there is a subset of individual structures within a population for which labelled data are available. These structures can be considered as candidate source domains. It then follows that individuals within a population for which data, or labels, are unavailable can be considered as target domains \cite{Gardner2021b}. Except in the most trivial scenarios, when using transfer learning there will be some differences between source and targets domains; in particular, it is typically assumed that the marginal distributions of observable data differ, i.e.\ $P(X_s) \neq P(X_t)$, and/or that the conditional distributions of labels differ, i.e.\ $P(\mathbf{y}_s|X_s) \neq P(\mathbf{y}_t|X_t)$. These discrepancies between source and target domain structures arise because of factors such as manufacturing variability; geometric, topological, and material differences; and operational and environmental differences.
	
	Several transfer-learning approaches have been applied to PBSHM. Various flavours of \textit{domain adaptation} have been used to harmonise source and target domains in the context of PBSHM  \cite{Gardner2018adaptation,Poole2022alignment,Gardner2022aircraft}. Other popular transfer learning techniques for PBSHM include neural approaches \cite{Yu2020adaptation,Soleimani2023zeroshot,Cao2018preprocessing} and multi-level modelling \cite{Bull2023hierarchical,Dardeno2023hierarchical}.
	
	\subsubsection{Negative Transfer}
	
	As alluded to earlier, an important consideration when applying transfer learning techniques in PBSHM is the possibility of negative transfer \cite{Wang2019negative}. Negative transfer is characterised by a degradation in predictive performance in the target domain following an information transfer attempt. Negative transfer is known to occur if the source joint distribution $P(X_s,\mathbf{y}_s)$ and target joint distribution $P(X_t, \mathbf{y}_t)$ are not sufficiently similar; or, if the algorithm used to conduct transfer is unable to find the correct mapping for other reasons, such as non-uniqueness of solutions.

	\subsection{Risk-informed Transfer Learning}\label{sec:risk}
	
	\subsubsection{Effects of Negative Transfer on Decision-making}
	
	Negative transfer has severe implications in the context of structural health monitoring and asset management. By definition, negative transfer results in erroneous classifications. In the most benign cases, these misclassifications may just be between two undamaged classes under different environmental conditions (e.g. normal and cold temperature). However, if the misclassifications occur between an undamaged class and a damage class, then at best an unnecessary action will be taken, and at worst a critical intervention will be missed leading to a catastrophic failure of a structure of interest. In both scenarios, the misclassifications result in some cost (or risk) being incurred when framed in the context of an operation and maintenance decision process. For this reason, it is discerning to anticipate and avoid negative transfer whenever possible.
	
	\subsubsection{Transfer Strategy Optimisation}\label{sec:optimisation}
	
	In order to avoid negative transfer and mitigate the potential impact on the operation and maintenance of target domain structures, a decision framework for optimising transfer learning strategies was proposed in \cite{Hughes2023decision}. This framework was based on a quantity termed \textit{value of information transfer}. For a more in-depth explanation of this decision framework the reader is directed to the original paper, for the purposes of the current paper a brief overview is provided here.
	
	Value of information (VoI) is a concept in decision theory defined to be the amount of resource (often money) a decision-maker should be willing to expend in order to gain access to information prior to making a decision. The concept of VoI has seen some application to traditional SHM in recent works \cite{Kamariotis2022,Hughes2022,Hughes2022c,Zhang2023quantifying}. In \cite{Hughes2023decision}, the notion of value of information was extended to PBSHM, and in particular to the information gained when applying transfer learning techniques. Termed the \textit{expected value of information transfer} (EVIT), this quantity can be interpreted as the price a decision-maker should be willing to pay in order to transfer information from a source domain prior to making decisions in a given target domain.
	
	As detailed in \cite{Hughes2023decision}, the EVIT can be computed as follows,
	
	\begin{equation}
		\text{EVIT}(\mathcal{T}) = \text{EU}(\mathcal{Q}|\mathcal{T}) - \text{EU}(\mathcal{Q}|\mathcal{T}_0) = P(\mathcal{Q}|\mathcal{T})\cdot U(\mathcal{Q}) - P(\mathcal{Q}|\mathcal{T}_0)\cdot U(\mathcal{Q}),
		\label{eq:EVIT}
	\end{equation}
	\noindent where $\text{EU}$ denotes the expected utility. $\mathcal{T}$ denotes a given transfer strategy parametrised by a source domain $\mathcal{D}_s$ and a transfer algorithm $\mathcal{A}$, with $\mathcal{T}_0$ corresponding to a null transfer strategy representing the case where no transfer is performed (i.e.\ $\mathcal{D}_s = \emptyset$ and $\mathcal{A} = \mathbb{I}$, where $ \mathbb{I}$ is an identity function). $\mathcal{Q}$ denotes some set of prediction quality criteria that can be related to utilities/costs in the context of a target-domain decision process via the utility function $U(\mathcal{Q})$. 
	
	Here, it is worth drawing attention to the fact that equation (\ref{eq:EVIT}) implies that $\text{EVIT}(\mathcal{T}_0) = 0$. Moreover, one can define negative and positive transfer from a decision-theoretic viewpoint -- simply that negative transfer is anticipated when $\text{EVIT}(\mathcal{T}) < 0$, and positive transfer is anticipated when $\text{EVIT}(\mathcal{T}) > 0$. Considering equation (\ref{eq:EVIT}) and the associated definition of negative transfer, it follows that one can minimise the risk associated with negative transfer via the optimisation,
	
	\begin{equation}
		\mathcal{T}^{\ast} = \mathcal{T}(\mathcal{D}_{s}^{\ast}, \mathcal{A}^{\ast}) = \underset{\mathcal{T}}{\text{argmax}} \big[ \text{EVIT}(\mathcal{T}) + U(\mathcal{T}) \big],
		\label{eq:T_opt}
	\end{equation}
	
	\noindent where an asterisk superscript denotes an optimal strategy/parameter, and $U(\mathcal{T})$ is a utility function specifying the cost associated with executing a given transfer strategy. Here, $\mathcal{T}^{\ast}$ comprises the source domain and transfer algorithm that minimise the risk of negative transfer.
	
\begin{figure}[h!]
		\centering
		\begin{tikzpicture}[x=1.7cm,y=1.8cm]
			
			% Nodes for plate GM
			\node[det] (UQ) {$U_{\mathcal{Q}}$} ;
			\node[latent,below=2.5cm of UQ] (Q) {$\mathcal{Q}$} ;
			\node[latent, right=1cm of Q] (S) {$\mathcal{S}$} ;
			\node[latent, above=1cm of S] (Rs) {$\mathcal{R}_s$} ;
			\node[obs, right=1cm of Rs] (R) {$\mathcal{R}_t$} ;
			\node[rectangle,draw=black,minimum width=0.8cm,minimum height=0.8cm,above=1cm of Rs] (T) {$\mathcal{T}$} ;
			\node[det, right=1cm of T] (UT) {$U_{\mathcal{T}}$} ;

			\edge {Q} {UQ} ; %
			\edge {T,S} {Q} ; %
			\edge {R,Rs} {S} ; %
			\edge {T} {UT,Rs} ; %

		\end{tikzpicture}
		\caption{An influence diagram representing a transfer decision process in terms of structural similarity between source and target domains \cite{Hughes2023decision}.}
		\label{fig:TransferID2}
	\end{figure}
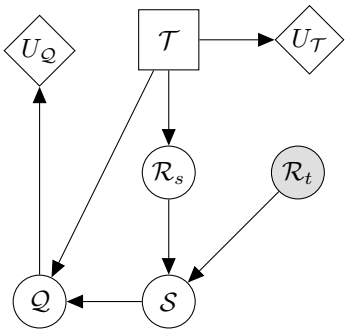

	In practice, to conduct the optimisation stated in equation (\ref{eq:T_opt}) without a computationally-expensive trial-and-error approach, one must forecast the post-transfer prediction qualities for target-domain data. Figure \ref{fig:TransferID2} shows an influence diagram that represents a factorisation of the objective function for the optimisation specified in equation (\ref{eq:T_opt}) that allows one to conduct the optimisation in a cost-effective manner. In essence, the approach leverages one of the key assumptions underpinning PBSHM; specifically, that members of a population that are structurally similar should yield improved transfer results in comparison to members of the population that are structurally disparate. For an in-depth explanation of the decision process presented in Figure \ref{fig:TransferID2} the reader is directed to \cite{Hughes2023decision}.
	
	In the factorisation presented in Figure \ref{fig:TransferID2}, the post-transfer prediction quality $\mathcal{Q}$ is assumed to be conditionally dependent on $\mathcal{S}$ -- the structural similarity between the source and targets domains. Additionally, $\mathcal{S}$ is conditionally dependent on $\mathcal{R}_s$ and $\mathcal{R}_t$, which denote the source domain and target domain expressed using a consistent representation (e.g.\ an attributed graph, or irreducible element model \cite{Brennan2021irreducible}). To summarise, the approach detailed in \cite{Hughes2023decision} requires a common representation of the source domain (specified in a transfer strategy $\mathcal{T}$) and target domain which allows the structural similarity to be assessed. This structural similarity can then be used to obtain a prediction of the post-transfer prediction quality using a pre-established mapping $P(\mathcal{Q}|\mathcal{S})$. Subsequently, these predictions can be used to evaluate the expected utility associated with the estimated prediction quality and thus compute the EVIT.
	
	The remainder of the current paper will focus on a numerical case study in order to demonstrate the process for establishing the crucial mapping $P(\mathcal{Q}|\mathcal{S})$ using the available source domains, the computation of the EVIT, and the optimisation of transfer strategies.
	
\section{Case Study: 10-DoF Systems}\label{sec:case_study}

\begin{figure}[ht!]
	\centering
	\includegraphics[scale=0.35]{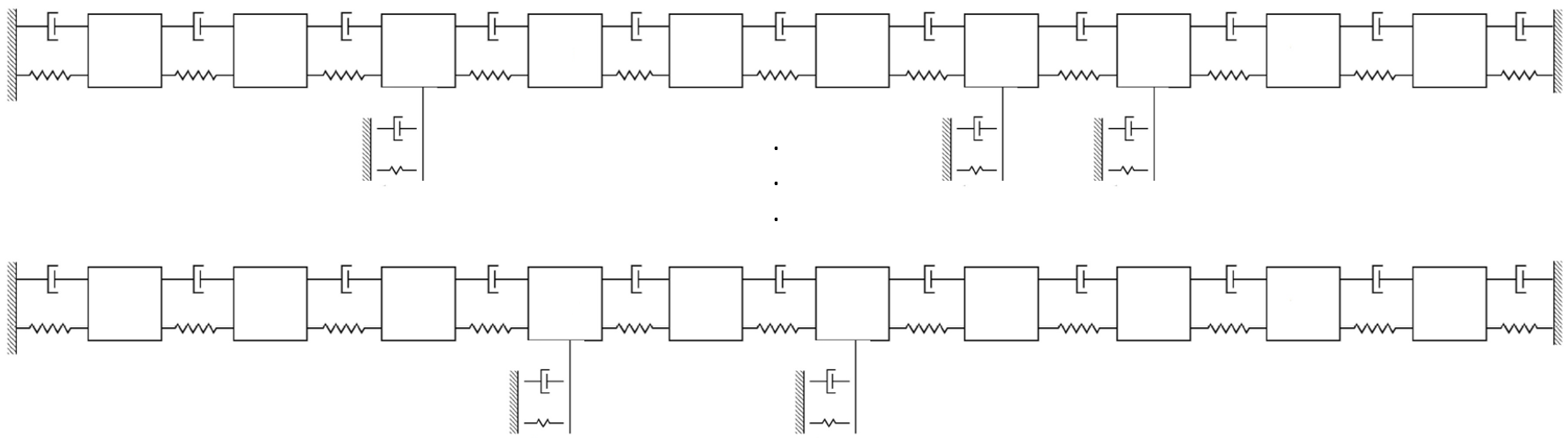}
	\caption{A sample of structures from a population of 10 degree-of-freedom systems with up to three random connections \cite{Poole2023negative}.}
	\label{fig:10dofs}
\end{figure}

To demonstrate the process for quantifying the EVIT and optimising transfer strategies, a numerical case study based on a population consisting of 20 10-degree-of-freedom (10-DoF) systems was considered -- this population was also the focus in \cite{Poole2023negative}. Examples of individual 10-DoF systems from the population are shown in Figure \ref{fig:10dofs}. 

Several sources of variability are present in the population. Each individual structure possesses one to three additional connections to ground at random locations across the six central masses. Additionally, variability amongst individuals arises via the assumptions that spring stiffnesses and damping coefficients are samples from a Gaussian distribution and a Gamma distribution, respectively. Although the systems in the population come from the same broader class of 10-DoF lumped-mass chains, because of the additional ground connections causing meaningful changes in the dynamics between individuals, the population as a whole can be considered weakly heterogeneous\footnote{Here, \textit{weakly} heterogeneous specifies that the degree of heterogeneity within the population at hand is less than the degree of heterogeneity that exists within, for example, a population consisting of both 10-DoF lumped-mass chains and helicopters.}.

Damage states for the structures were simulated by reducing the stiffness of the first 10 springs, on an individual basis, by 50\%. To ensure that the label spaces $Y$ were consistent between individual structures, the added connections to ground were not considered to be damaged. Natural frequencies and modal matrices for each member of the population were obtained from the system parameters via eigendecomposition. Mode shapes were subsequently normalised to unit length. To generate a labelled dataset for each structure, 250 data points from the undamaged class were sampled in addition to 25 data points for each of the damage states, resulting in an overall dataset size of 500 points with an even split between damaged and undamaged data.

Using the described population and associated data, one can construct a training dataset that can be used to establish a mapping between structural similarity and post-transfer prediction quality by generating transfer tasks between pairs of members within the population. This process involves obscuring the labels for one domain in the pair (akin to a pseudo-target domain), while using the other domain as a source domain. Following each transfer, the target domain prediction quality can be assessed.

Here, details of the structural similarity measure, classification task, and transfer tasks used in the current case study are provided.

\subsection{Classification Task}

The target-domain classification task used in the current case study is to predict the ten damage-state labels $y \in \{1,...,10\}$ (including the undamaged state), using the natural frequencies of the system $\mathbf{x} \in \mathbb{R}^{10}_{+}$ as features. To achieve the mapping $f: X \rightarrow Y$, a $k$-nearest-neighbour algorithm was used with $k=1$.

To evaluate predictive performance on the classification task, the following prediction quality measures were considered:

\begin{itemize}
	\item True Prediction Rate (TR): $\text{TR} = \frac{\# \text{ of true predictions}}{\# \text{ of predictions}}$,
	\item False Positive Prediction Rate (FPR): $\text{FPR} = \frac{\# \text{ of false positive predictions}}{\# \text{ of predictions}}$,
	\item False Negative Prediction Rate (FNR): $\text{FNR} = \frac{\# \text{ of false negative predictions}}{\# \text{ of predictions}}$,
	
\end{itemize}

\noindent where true predictions correspond to instances where the correct label is predicted, false positives (or type-I errors) correspond to instances where the predicted label is a specific damage label; however, the true label is a differing damage condition or undamaged. False negatives (or type-II errors) are instances where the predicted label corresponds to undamaged, when in reality the correct label is one of the damage states. For a more convenient notation, these prediction quality measures are concatenated into a vector $\mathbf{q} = \{\text{TR},\text{FPR},\text{FNR}\}$.

These prediction quality measures were selected as they are agnostic to the size of the target-domain dataset but also because they can be easily interpreted in the context of SHM decision-making. In an operation and maintenance decision process such as inspection and repair planning, true predictions would correspond to the structure being kept operational as a correct course of action could be selected; in terms of utility, this type of prediction would correspond to some nominal positive value. False-positive predictions, where damage is predicted when in fact a structure is undamaged, will cause unnecessary inspections, and/or unnecessary reapir/replacements of components, in the context of operation and maintenance decision processes. In terms of utility, these types of predictions will have a cost, or negative value, associated. False-negative predictions are more concerning than the other types of prediction. In the context of SHM, this type of prediction means damage has not been identified by the monitoring system, meaning erroneous actions will be taken. Specifically, it means that no intervention would be made to mitigate the effects of the existing damage, thereby compromising the safety/integrity of the structure. In terms of utility, this type of error will typically have a large associated negative utility, corresponding to the risk related the failure of the structure which would including both the financial cost of losing an asset and (potentially) the threat to human life caused by structural failure.

For the purposes of the current case study, these translations from prediction quality to decision consequences are encoded in the utility function provided in Table \ref{tab:UQ}.

\begin{table}[h!]
	\centering
	\caption{The utilities associated with the prediction types True, False Positive, and False Negative.}
	\begin{tabular}{lllll}
		\cline{1-4}
		\multicolumn{1}{|l|}{\textbf{Prediction Type}} & \multicolumn{1}{c|}{True} & \multicolumn{1}{c|}{False Positive} & \multicolumn{1}{c|}{False Negative} &  \\ \cline{1-4}
		\multicolumn{1}{|l|}{\textbf{Utility}} & \multicolumn{1}{c|}{5} & \multicolumn{1}{c|}{-10} & \multicolumn{1}{c|}{-50} &  \\ \cline{1-4}
	\end{tabular}
	\label{tab:UQ}
\end{table}
 
 \subsection{Transfer Tasks}\label{sec:tasks}
 
 In general, the number of transfer tasks that can be constructed from a set of source domains is given by $N_T = N_s^2 - N_s$ since each structure can be used as a source domain while each of the remaining structures can be considered as a target domain. This equality means that from the population of 20 10-DoF systems, one can construct 380 unique transfer tasks and thus one has 380 data points from which to learn a mapping from similarity to prediction quality.
 
To accomplish information transfer for each task, a simple but highly-effective form of domain adaptation known as normal condition alignment (NCA) was applied; a detailed explanation of this approach can be found in \cite{Poole2022alignment}. In NCA, data are translated and scaled such that the undamaged data in the target domain aligns with the undamaged data in the source domain. Assuming source domain data are standardised, NCA is given by,

\begin{equation}
	z_t^{(j)} = \bigg(\frac{x_t^{(j)} - \mu_{t,n}}{\sigma_{t,n}}\bigg) \sigma_{s,n} + \mu_{s,n}
\end{equation}
 
\noindent where $z_t$ are the transformed target-domain features, $x_t$ are the original target-domain features, and $\mu_{t,n}$, $\sigma_{t,n}$, $\mu_{s,n}$, and $\sigma_{s,n}$ are the mean and standard deviation of the normal-condition data for the target and source domains, respectively. Following the application of NCA, predictions can be made for data in the target domain as the kNN classifier can leverage labelled data from the source domain. The NCA approach used here assumes that there are labelled data available corresponding to the structure being in an undamaged condition; this is not an entirely unreasonable assumption to make as it is typically believed that at the beginning of a monitoring campaign, the structure is in an undamaged/normal condition. This assumption would not necessarily hold if a monitoring system has been retrofitted.

Because the labels for each pseudo-target domains and were only obscured for the purposes of generating transfer tasks and are in fact available, the post-transfer prediction quality for each task can be assessed by comparing the predicted values with the ground-truth targets. This process generated the data $Q = \{ \mathbf{q}_n \}_{n=1}^{N_T}$ to be used as the targets for establishing the mapping between structural similarity and prediction quality.

Here, it is worth stating that to simplify the analysis somewhat, in the current case study only one transfer algorithm is considered. The process described thus far can be generalised to consider multiple transfer algorithms simply by repeating the process for each unique algorithm. Furthermore, it is assumed that the cost of transfer is invariant across the target domains, thus it can be ignored  for the decision analyses in the current case study.

\subsection{Structural Similarity}\label{sec:mac}

As shown in \cite{Poole2023negative}, structural similarity can be used as a predictor of post-transfer prediction quality. For the case study in hand, it was opted to follow the approach used in \cite{Poole2023negative}. This approach involves using the mode shapes of the structures in the population as a representation that can be used to assess structural similarity in a physics-informed manner. In particular, the modal assurance criterion (MAC) provides a similarity score between mode shapes that can be used as a proxy for structural similarity. The MAC is calculated as follows,

\begin{equation}
	\text{MAC}(i,j) = \frac{|\bm{\phi}_{s,i}^{\top}\bm{\phi}_{t,j}|^2}{\bm{\phi}_{s,i}\bm{\phi}_{s,i}^{\top}\bm{\phi}_{t,j}^{\top}\bm{\phi}_{t,j}}
\end{equation}

\noindent where $\bm{\phi}_{s,i}$ denotes the $i^{\text{th}}$ modal vector from the source-domain modal matrix $\Phi_s$, $\bm{\phi}_{t,j}$ denotes the $j^{\text{th}}$ modal vector from the target-domain modal matrix $\Phi_t$, and $\text{MAC}(i,j) \in [0,1] $, with 0 indicating no correspondence between modes, and 1 indicating total correspondence between modes.

If the modal matrices for both the source and target domains contain similar modal vectors and in the same order, then values in the leading diagonal of the MAC matrix should be close to unity. However, in general, there is no guarantee that the corresponding modes will appear in the same order between the source and target domains, so the vectors of the MAC matrix are permuted such that the largest entries lie on the diagonal. From the permuted MAC matrix, one can define a proxy for structural similarity as follows,

\begin{eqnarray}
	\varsigma(\text{MAC}) = \frac{\trace{\text{MAC}}}{N_m}
\end{eqnarray}

\noindent where $\trace{\cdot}$ denotes the trace of a matrix, and $N_m$ is the number of modes considered, and is used to normalise the similarity measure such that $\varsigma \in [0,1]$.

For each of the transfer tasks, the MAC-based similarity measure was computed for the source and target domain. This process generated the data $\bm{\varsigma} = \{ \varsigma_n \}_{n=1}^{N_T}$ to be used as a predictive feature when establishing a mapping from structural similarity to prediction quality.
 
 \subsection{Forecasting Prediction Quality}\label{sec:forecast}
 
 In the transfer strategy decision framework being used for the current case study,  to compute the EVIT, a probabilistic mapping $g : \mathcal{S} \rightarrow \mathcal{Q}$ is required. Via the processes detailed in Sections \ref{sec:tasks} and \ref{sec:mac}, a dataset $\mathcal{D}_T = \{ \varsigma_n , \mathbf{q}_n \}_{n=1}^{N_T} = \{ \bm{\varsigma}, Q \}$ was generated. With $\varsigma \in \mathcal{S}$ and $\mathbf{q} \in \mathcal{Q}$, the data $\mathcal{D}_T$ were used to learn the mapping $g$.
 
 In learning a mapping $g$, one is essentially performing a probabilistic vector-valued regression. In addition to the regression being vector-valued, there were several constraints that were to be adhered to; specifically:
 
 \begin{itemize}
 	\item $0 \leq \text{TR},\text{FPR}, \text{FNR} \leq 1$
 	\item $\text{TR}+\text{FPR}+\text{FNR} = 1$.
 \end{itemize}
 
 These properties of the prediction quality measures essentially mean that $\mathcal{Q}$ is a 2-simplex. Given that $\mathcal{Q}$ is a 2-simplex, and probabilistic predictions were required, it was natural to assume that the $\mathbf{q}_n$ were Dirichlet distributed,
 
 \begin{equation}
 	\mathbf{q}_n \sim \text{Dir}(\bm{\alpha}_n)
 \end{equation}
 
 \noindent where $\bm{\alpha}_n = \{\alpha_{1},\alpha_{2},\alpha_{3} \}$ with $\alpha_k > 0$, and where $\alpha_1$, $\alpha_2$, $\alpha_3$ are the concentration parameters for TR, FPR, and FNR, respectively. By making this assumption, learning the mapping $g$ amounted to learning a function to regress from the structural-similarity measures $\varsigma_n$ to the latent concentration parameters of the Dirichlet distribution $\bm{\alpha}_n$. To perform this vector-valued regression, the authors opted to use an artificial neural network -- specifically, a multi-layer perceptron (MLP). The architecture of the MLP used is provided in Table \ref{tab:mlp}.
 
 \begin{table}[h!]
 	\centering
 	\caption{The architecture of the multi-layer perceptron.}
 	\begin{tabular}{llllll}
 		\cline{1-5}
 		\multicolumn{1}{|l|}{\textbf{Layer}} & \multicolumn{1}{c|}{Input} & \multicolumn{1}{c|}{Hidden 1} & \multicolumn{1}{c|}{Hidden 2} & \multicolumn{1}{c|}{Output} & \\ \cline{1-5}
 		\multicolumn{1}{|l|}{\textbf{Units}} & \multicolumn{1}{c|}{1} & \multicolumn{1}{c|}{8} & \multicolumn{1}{c|}{12} & \multicolumn{1}{c|}{3} &  \\ \cline{1-5}
 		\multicolumn{1}{|l|}{\textbf{Activation}} & \multicolumn{1}{c|}{\texttt{softplus}} & \multicolumn{1}{c|}{\texttt{softplus}} & \multicolumn{1}{c|}{\texttt{softplus}} & \multicolumn{1}{c|}{\texttt{softplus}} &  \\ \cline{1-5}
 	\end{tabular}
 	\label{tab:mlp}
 \end{table}
 
From Table \ref{tab:mlp} it can be seen that \texttt{softplus} activation functions were used, this was to ensure that the concentration parameters were positive.

The loss function used to evaluate the performance of the MLP expressed two distinct objectives, the first component $\ell_1$ corresponded to the negative log-marginal-likelihood for the Dirichlet distribution, given by,

\begin{equation}
	\ell_{1,n} = - \log\Gamma(\alpha_{0,n}) + \sum_{k=1}^{3}\log\Gamma(\alpha_{k,n}) - \sum_{k=1}^{3}(\alpha_{k,n} - 1)\cdot \log q_{k,n}
\end{equation} 

\noindent where $\Gamma$ denotes the gamma function and $\alpha_0 = \sum_{1}^{3}\alpha_k$. Here, $q_1$, $q_2$, and $q_3$ correspond to observations of TR, FPR, and FNR, respectively. Including this component in the loss function means models that maximise the likelihood of the data were preferred.

In addition to the Dirichlet negative log-marginal-likelihood component of the loss function, an additional component was introduced such that monotonically-increasing functions were preferred for modelling TR. This property was desirable as it reflects the belief that applying transfer learning to more similar structures should yield better outcomes than when applying it to more dissimilar structures. This component of the loss function was defined as follows,

\begin{equation}
	\ell_{2,n} = 
	\begin{cases}
		-\lambda, & \frac{\alpha_{1,n}}{\alpha_{0,n}} < \frac{\alpha_{1,n-1}}{\alpha_{0,n-1}}, \\
		0, & \text{otherwise},
	\end{cases}
\end{equation}

\noindent where $\lambda$ is a constant that controls the strength of $\ell_2$ in the loss function, relative to the Dirichlet log-marginal-likelihood component -- i.e.\ larger values of $\lambda$ induce a stronger preference for monotonic functions. For the current case study, $\lambda=1$.

Finally, an overall loss function $\mathcal{L}$, considering all available training data, was constructed by combining $\ell_1$ and $\ell_2$ as follows,

\begin{equation}
	\mathcal{L} = \frac{1}{N_T}\sum_{n=1}^{N_T}(\ell_{1,n} + \ell_{2,n}).
\end{equation}

To summarise, the methodology detailed in the current section was applied to a numerical case study formulated around a population consisting of 20 10-degree-of-freedom systems. In the next section, the results of this case study will be presented.
 
\section{Results}\label{sec:results}

\begin{figure}[h!]
	\centering
	\includegraphics[width=0.45\linewidth]{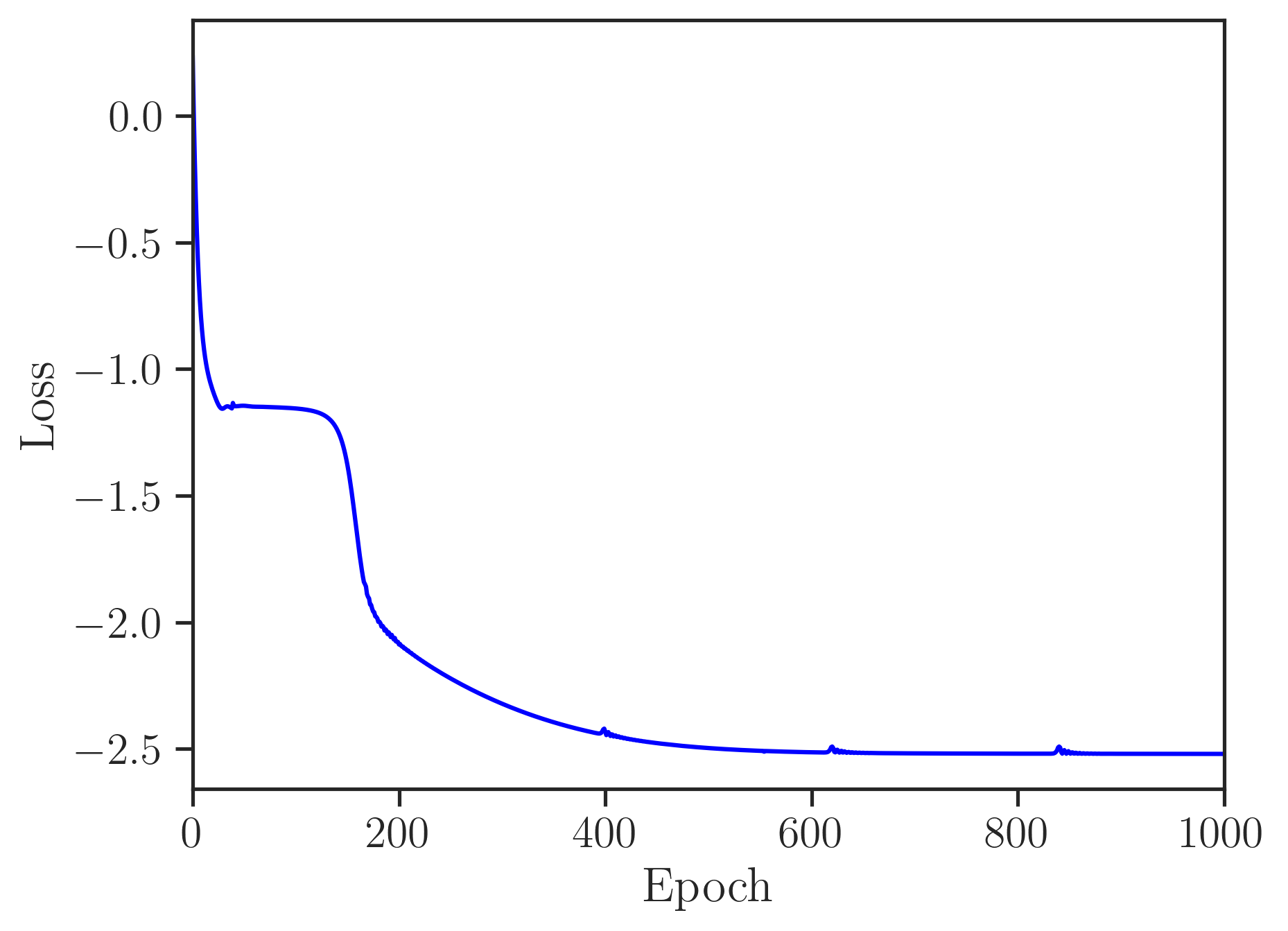}
	\caption{The loss plotted against the number of epochs during the training of the MLP.}
	\label{fig:loss}
\end{figure}

The methodology provided in the previous section was applied on the case study based on 10-DoF systems. Before the EVIT could be assessed, the artificial neural network described in Section \ref{sec:forecast} was trained using all available data in $\mathcal{D}_T$ via the Adam optimiser \cite{Kingma2014adam} -- a stochastic gradient-descent method. The training loss over 1000 epochs is shown in Figure \ref{fig:loss}.

It can be seen from Figure \ref{fig:loss} that after approximately 600 epochs the optimiser is able to converge on an optimum. The initial plateau at around 100 epochs indicates that there are local minima in the parameter space for the given loss function. The blips in the loss that occur after convergence can be attributed to the stochastic nature of the optimisation algorithm. 

\begin{figure}[h!]
	\centering
	\begin{subfigure}[b]{.45\linewidth}
		\includegraphics[width=\linewidth]{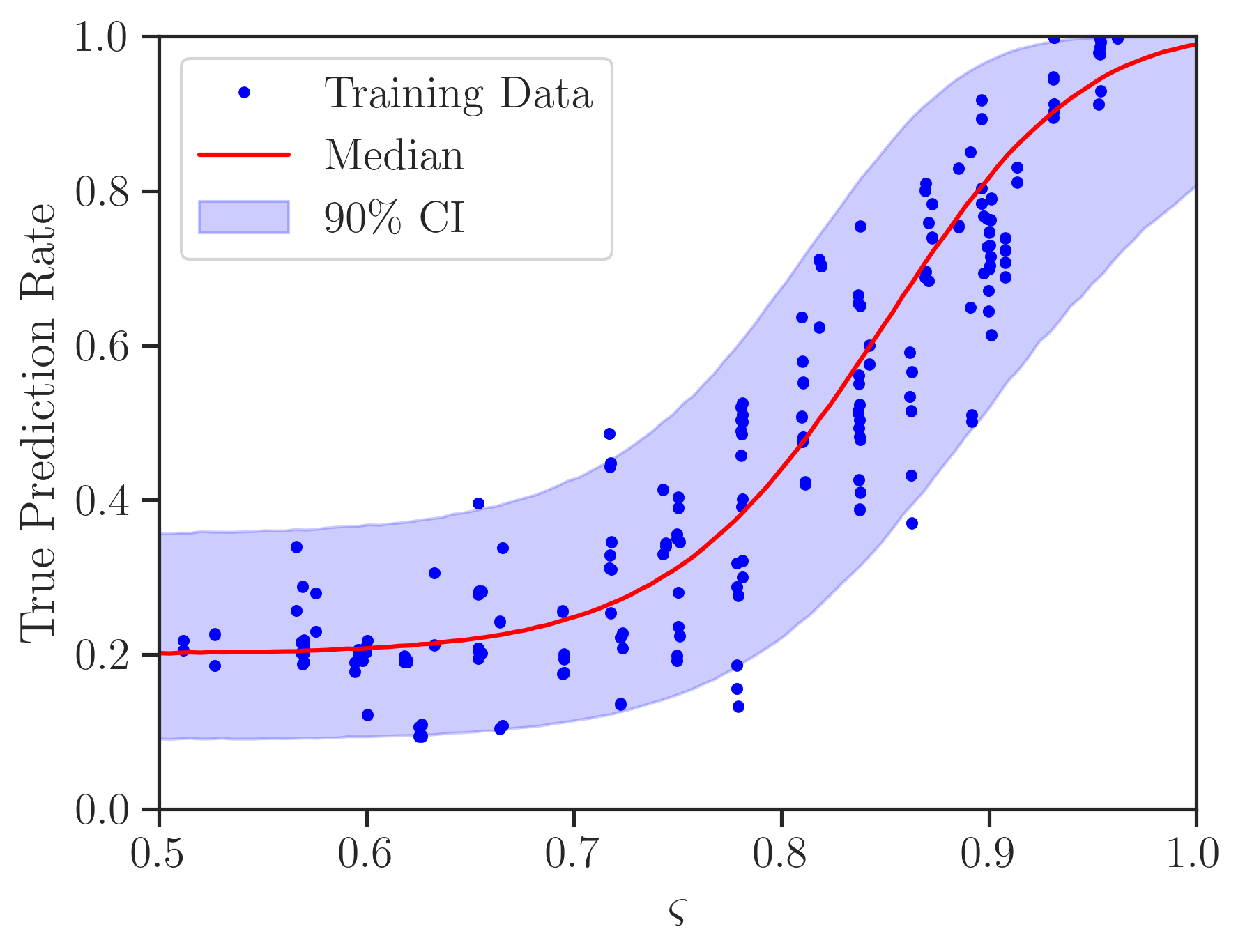}
		\setcounter{subfigure}{0}%
		\caption{}\label{fig:TR}
	\end{subfigure}
	
	\begin{subfigure}[b]{.45\linewidth}
		\includegraphics[width=\linewidth]{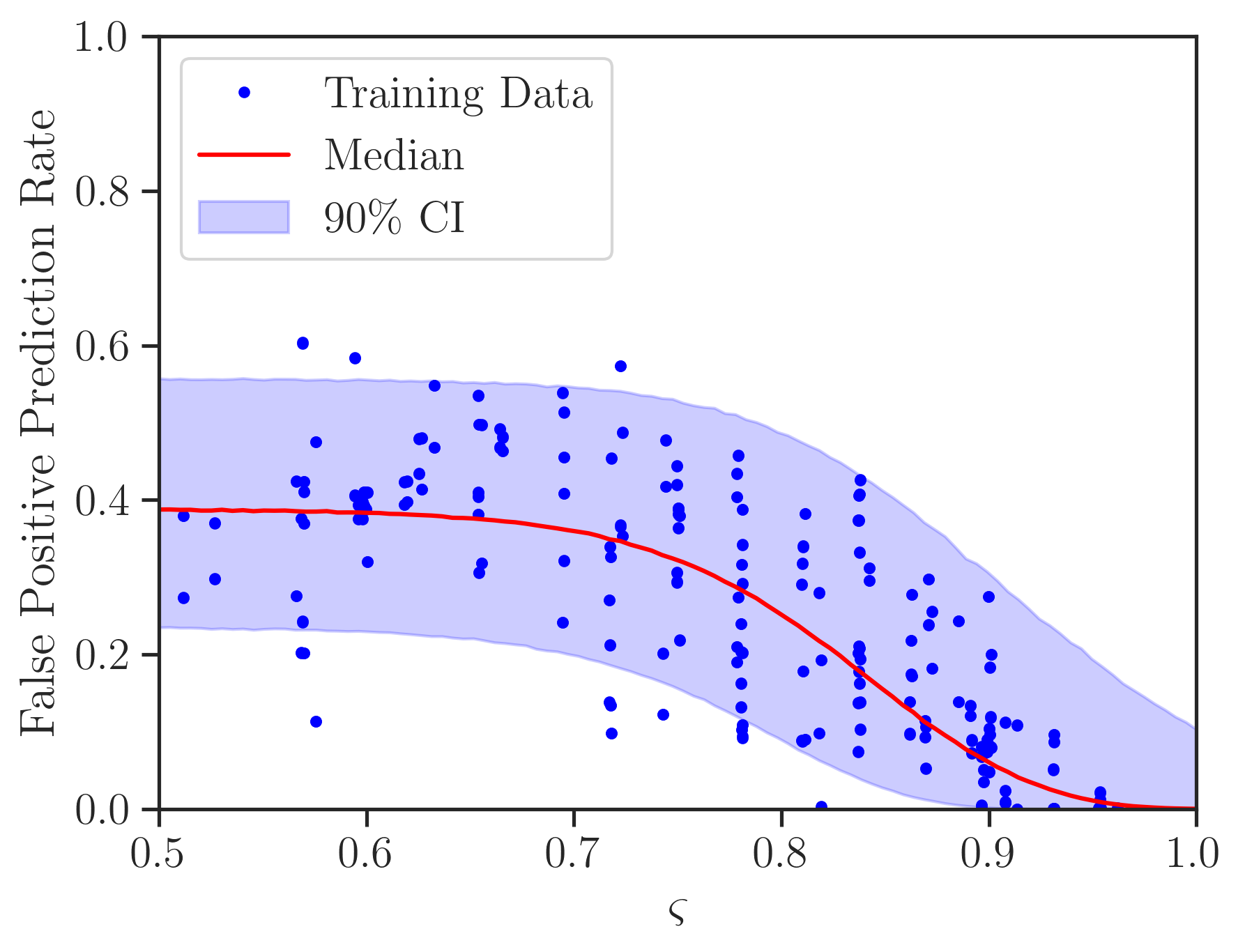}
		\caption{}\label{fig:FPR}
	\end{subfigure}
	\begin{subfigure}[b]{.45\linewidth}
		\includegraphics[width=\linewidth]{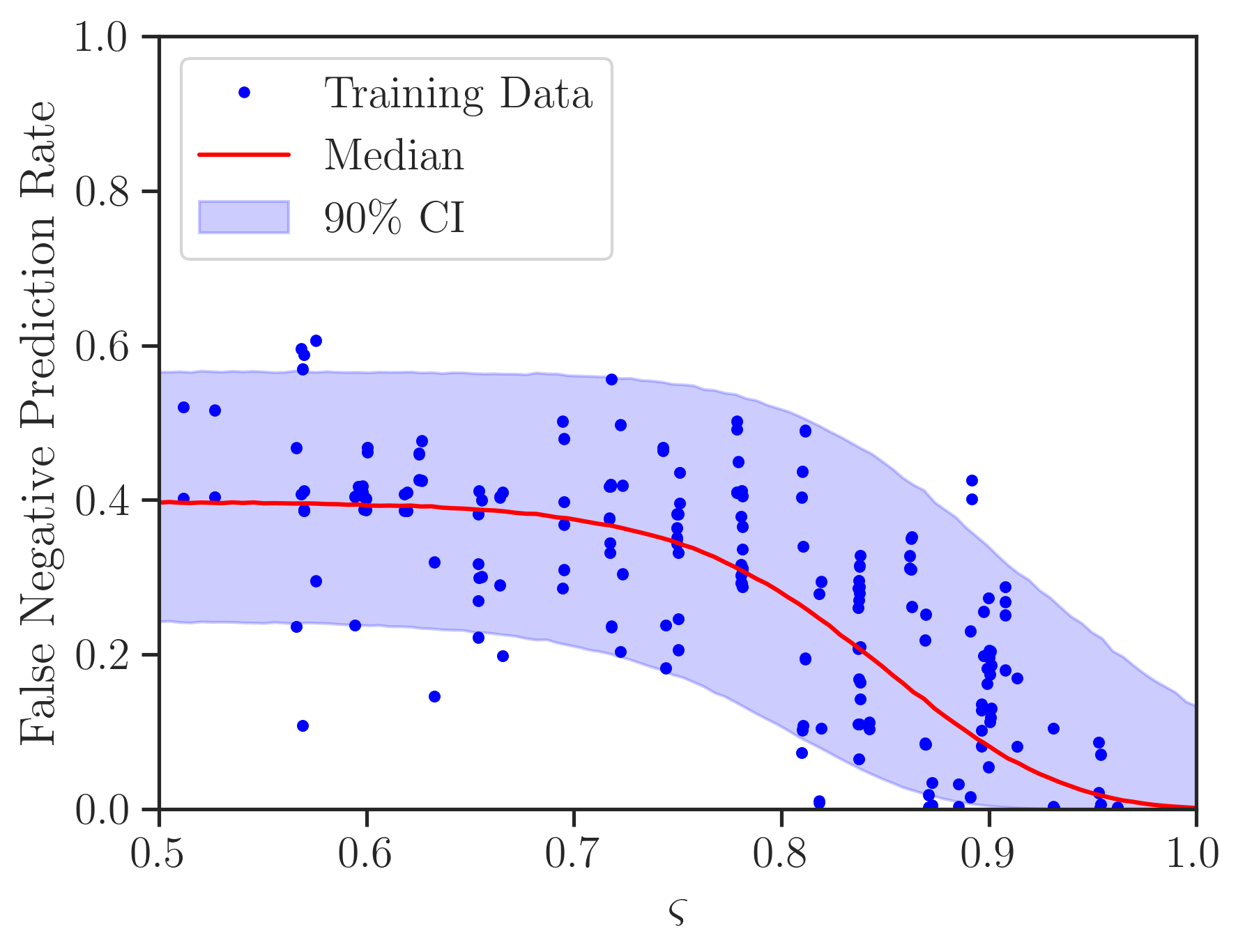}
		\caption{}\label{fig:FNR}
	\end{subfigure}
	\caption{Learned probabilistic functions $p(q_k|\varsigma)$ for (a) the true prediction rate, (b) the false-positive prediction rate, and (c) the false-negative prediction rate. Blue shaded regions indicate the 90\% confidence intervals (CI) computed using samples taken from the Dirichlet distributions.}
	\label{fig:results}
\end{figure}

Figure \ref{fig:results} shows the results of the vector-valued regression detailed in Section \ref{sec:forecast}. Specifically, it shows probabilistic functions $p(q_k|\varsigma)$ for TR, FPR, and FNR, that constitute the mapping $g$.

It can be seen from Figure \ref{fig:TR} that there is strong positive correlation between the proxy for structural similarity and the post-transfer true prediction rate. This result is extremely promising as it indicates that, should a sufficiently-similar source domain be available, predictions can almost certainly be improved by conducting transfer learning. In a similar vein, both Figure \ref{fig:FPR} and Figure \ref{fig:FNR} demonstrate negative correlation between the false predictions and structural similarity; this result is of course expected when considering the results in Figure \ref{fig:TR}. By observing that in Figure \ref{fig:TR}, for high similarities $\varsigma > 0.9$, the median lies closer to the upper bound of the 90\% confidence interval (CI) -- one can infer from this that for these values of similarity, the probability distribution is skewed towards larger values of TR; this is a favourable result. From Figure \ref{fig:results} it can be seen that the median functions learned are monotonic; the predictions for TR are monotonically-increasing with $\varsigma$, and the predictions for FPR and FNR are monotonically-decreasing with $\varsigma$. This result shows that the functions learned align with the belief that structural similarity improves the outcomes of information-transfer.

\begin{figure}[h!]
	\centering
	\includegraphics[width=0.45\linewidth]{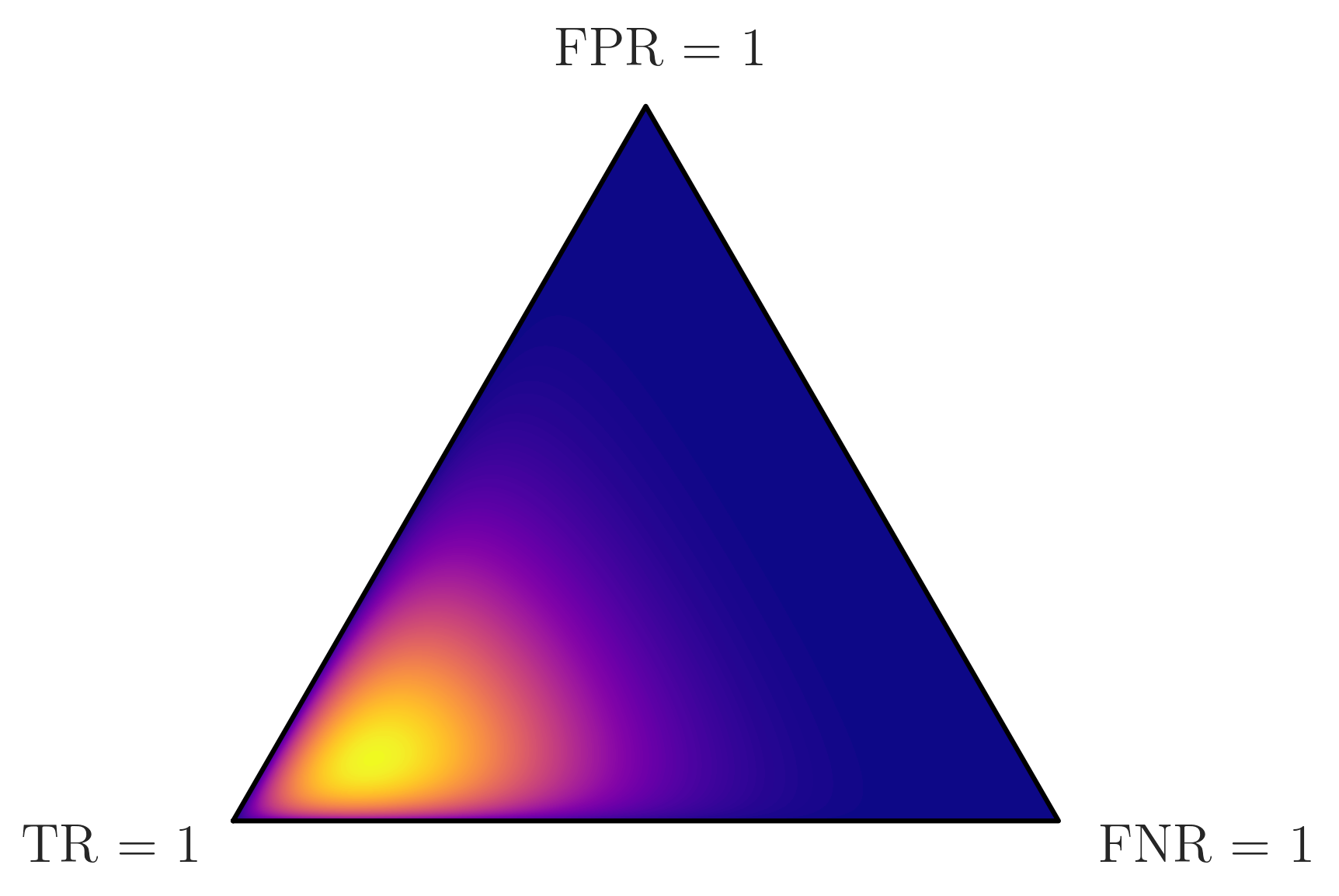}
	\caption{The Dirichlet probability density over the 2-simplex formed by the prediction quality measures TR, FPR, FNR for $\varsigma = 0.85$. Yellow/orange indicates regions of high probability density, whereas blue indicates regions of low probability density.}
	\label{fig:simplex}
\end{figure}

With the probabilistic mapping established between structural similarity and post-transfer prediction quality, one can now imagine a situation where a new target-domain structure is introduced to the population. In this event, using the available undamaged/normal condition data for the target domain, the mode shapes can be estimatd and compared with those for each of the available source domains. For illustrative purposes, suppose that one finds that the most similar source domain has $\varsigma=0.85$; this similarity score can be used an an input to the MLP to obtain concentration parameters. These concentration parameters would in turn specify the Dirichlet probability density over the post-transfer prediction quality. Figure \ref{fig:simplex} shows the probability denisity function over the prediction quality 2-simplex for $\varsigma=0.85$ as a heatmap where the yellow and orange indicate regions of high probability density and dark blue indicates regions of low probability density.

From Figure \ref{fig:simplex}, one can see that a significant portion of the probability mass lies close to the vertex representing $\text{TR} = 1$. Naturally, this result is in correspondence with those shown in Figure \ref{fig:results} as the simplex-view represents the same information, albeit in a more compact format, as the predictions in Figure \ref{fig:results} for a slice taken at $\varsigma = 0.85$. A distribution such as the one shown in Figure \ref{fig:simplex} can be sampled from to generate vectors corresponding to possible post-transfer prediction quality measures -- this is useful for evaluating the EVIT.

Suppose now that the target-domain dataset comprises $M=200$ unlabelled observations. By taking the product of samples of the prediction rates from the Dirichlet distribution with $M$, one obtains a distribution over the number of true, false-positive, and false-negative predictions. Taking the product of this distribution of the number of each prediction type with the utility function given in Table \ref{tab:UQ} yields a probability distribution over the utility associated with each prediction type. Taking the expectation of this distribution, and summing over the prediction types yields the expected utility $\text{EU}(\mathcal{Q}|\mathcal{T})$ -- one of the terms required to calculate the EVIT per equation (\ref{eq:EVIT}). The remaining term, $\text{EU}(\mathcal{Q}|\mathcal{T}_0)$, corresponds the the expected utility associated with the prediction-quality measures following the application of a null transfer strategy, i.e.\ following no transfer. For simplicity, in the current case study, for the case of performing no transfer, the numbers of each prediction-type for the target domain were assumed to be uniformly distributed, representing the random allocation of unknown labels. This assumption gave a constant value of $\text{EU}(\mathcal{Q}|\mathcal{T}_0)=-3666.67$.

\begin{figure}[h!]
	\centering
	\includegraphics[width=0.45\linewidth]{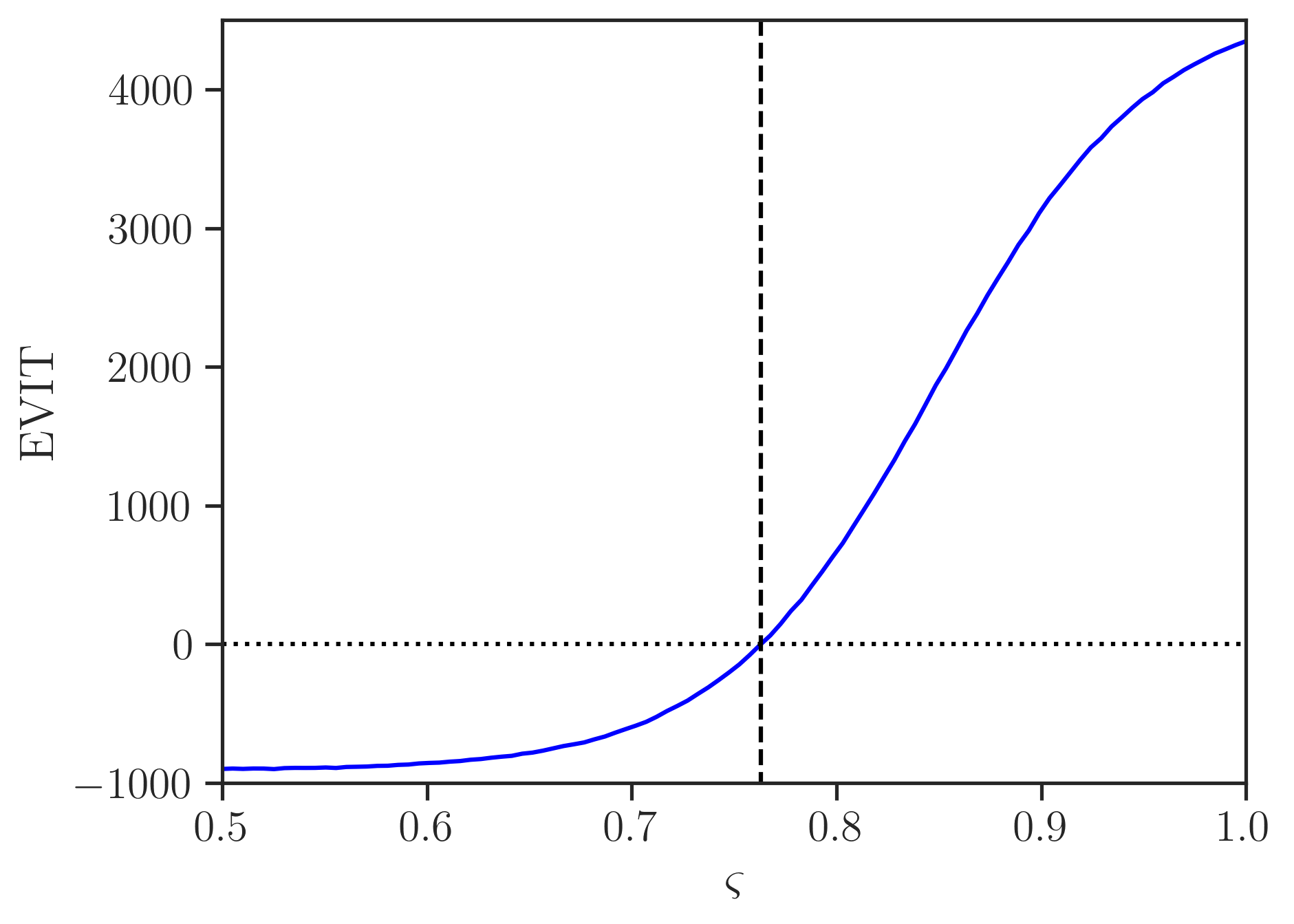}
	\caption{The expected value of information transfer as a function of (proxy) structural similarity. The dotted line marks the $\text{EVIT}=0$ threshold for positive transfer, and the dashed line marks the structural similarity $\varsigma=0.76$ beyond which positive transfer is achieved.}
	\label{fig:EVIT}
\end{figure}

By computing $\text{EU}(\mathcal{Q},\mathcal{T})$, using the methodology described above, for a range of transfer strategies corresponding to performing transfer learning with source domains over a range of $\varsigma$, one can obtain the EVIT as a function of structural similarity. For the case study used in the current paper, this function is shown in Figure \ref{fig:EVIT}.

Figure \ref{fig:EVIT} shows clearly that EVIT is a strictly-increasing function of $\varsigma$ -- this result is very positive but again expected when considered in the context of the predictions shown in Figure \ref{fig:results}. Using this property of the function shown in Figure \ref{fig:EVIT}, one can develop a heuristic for selecting the optimal transfer strategy $\mathcal{T}^{\ast}$. For the function shown, the optimal transfer strategy can be selected simply by selecting the source domain that possesses the highest structural similarity with the target domain\footnote{This heuristic assumes that only single-source transfer strategies are considered.}. Furthermore, recalling the decision-theoretic definition of negative transfer given in Section \ref{sec:optimisation}, one can use Figure \ref{fig:EVIT} to evaluate the minimum structural similarity at which positive transfer is expected by finding the value of $\varsigma$ at which the criterion $\text{EVIT} > 0$ is met. For the current case study, the threshold is crossed when $\varsigma = 0.76$. The threshold, and the value at which the threshold is crossed, are denoted by dotted and dashed lines, respectively.

To summarise, for the case study detailed in Section \ref{sec:case_study}, monotonic and probabilistic functions were learnt that map from the structural similarity between source and target structures to post-transfer prediction quality measures. Subsequently, these mappings were used to estimate the number of prediction types for a given target-domain dataset size, and the associated expected utility. This information was used to determine how the expected value of information transfer varies as the structural similarity between the source and target domain increases. It was found that EVIT montonically-increases with structural similarity. This knowledge can be used to optimise transfer strategies by selecting source domains that maximise the expected value of information transfer. This process is equivalent to minimising the risk of negative transfer. In the next section, a few considerations and limitations are discussed.

\section{Discussion}\label{sec:Discussion}
	
While it was shown in the previous section that transfer strategies can be optimised such that the risk of negative transfer is minimised, the methodology detailed is not without its limitations. Possibly the most significant drawback of the method is that it becomes infeasible to establish a mapping between similarity and prediction quality if there are few candidate source domains, as the data will be too sparse. The scarcity of the data will either prevent a function from being learnt, or result in predictions with very high uncertainty, thereby reducing the estimate of EVIT. The silver lining to this cloud is that the number of datapoints available for learning increases quadratically with the number of candidate source domains. As the number of source domains increases, and the size of dataset $\mathcal{D}_T$ grows, functions will be more readily learnable and quantification of uncertainty will improve. Furthermore, when it is deemed that a sufficient amount of data is available for training, the data can be partitioned into training and validation datasets -- using a validation dataset will provide resilience to overfitting when learning the required mappings, additionally it will allow for the optimisation of hyperparameters such as those specifying the neural network architecture in Table \ref{tab:mlp}.

A further drawback of the overall methodology outlined in Section \ref{sec:optimisation}, is that the size of the decision space grows rapidly when considering transfer strategies that involve transferring information from multiple source domains and/or multiple different transfer-learning algorithms. In the context of the case study presented in Sections \ref{sec:case_study} and \ref{sec:results}, this issue would require a separate mapping $g$ for each of the candidate transfer-learning algorithms and would greatly increase the computational cost of conducting the optimisation.

Despite some drawbacks, the approach outlined has the advantage of being quite generalisable. For example, the representation and similarity measure are not limited to being the mode shapes and a metric based on the MAC. Instead, it would be entirely feasible to represent the candidate source domains and target domains graphically, such as via an IE model and use a graph-based distance, or graph matching network, to obtain structural similarities \cite{Brennan2022quantifying}. In fact, this approach would yield the advantage that the EVIT could be evaluated, and transfer strategy optimised, without the need for \textit{any} target domain data. Demonstrating this advantage is left as future work.

An additional avenue for future work is to show the quantification of expected value of information transfer using a population of experimental structures. In the final section of the paper, concluding statements are provided.

	\section{Conclusions}\label{sec:conclusions}
	
	To conclude, although traditional approaches to SHM are troubled by issues pertaining to data scarcity, PBSHM offers a general framework for overcoming such issues. PBSHM achieves this advantage via technologies that allow for the transfer of information between individual structures within a population. When conducting information transfer, there is a risk of degrading prediction performance; a phenomenon known as negative transfer. In practice, negative transfer has a meaningful cost in terms of finances and/or safety, as incorrect predictions lead to erroneous O\&M decisions being made which undermine the integrity/performance of the structures. The current paper demonstrates that transfer-learning strategies can be optimised to minimise the risk associated with negative transfer. This optimisation is accomplished by considering the expected value of information transfer which can be estimated by considering information transfers between the existing available source domains. The results presented in the current paper provide exciting motivation for the pursuit of population-based SHM technologies.
	
	\section*{Acknowledgements}
	The authors would like to gratefully acknowledge the support of the UK Engineering and Physical Sciences Research Council (EPSRC) via grant reference EP/W005816/1. For the purposes of open access, the authors have applied a Creative Commons Attribution (CC BY) license to any Author Accepted Manuscript version arising.
	
	\bibliographystyle{unsrt}
	\bibliography{IMAC2022}

\begin{thebibliography}{10}

\bibitem{Farrar2013}
C.R.\ Farrar and K.\ Worden.
\newblock {\em {Structural Health Monitoring: A Machine Learning Perspective}}.
\newblock John Wiley {\&} Sons, Ltd, United Kingdom, 2013.

\bibitem{Rytter1993}
A.\ Rytter.
\newblock {\em {Vibration Based Inspection of Civil Engineering Structures}}.
\newblock {Ph.D. Thesis}, Aalborg University, 1993.

\bibitem{Papakonstantinou2014a}
K.G.\ Papakonstantinou and M.\ Shinozuka.
\newblock {Planning structural inspection and maintenance policies via dynamic
  programming and Markov processes. Part I: Theory}.
\newblock {\em Reliability Engineering \& System Safety}, 130:202--213, 2014.

\bibitem{Schobi2016}
R.\ Sch{\"{o}}bi and E.N.\ Chatzi.
\newblock {Maintenance planning using continuous-state partially observable
  Markov decision processes and non-linear action models processes and
  non-linear action models}.
\newblock {\em Structure and Infrastructure Engineering}, 12(8):977--994, 2016.

\bibitem{Vega2020a}
M.A.\ Vega and M.D.\ Todd.
\newblock {A variational Bayesian neural network for structural health
  monitoring and cost-informed decision-making in miter gates}.
\newblock {\em Structural Health Monitoring}, page 1475921720904543, 2020.

\bibitem{Hughes2021}
A.J.\ Hughes, R.J.\ Barthorpe, N.\ Dervilis, C.R.\ Farrar, and K.\ Worden.
\newblock {A probabilistic risk-based decision framework for structural health
  monitoring}.
\newblock {\em Mechanical Systems and Signal Processing}, 150:107339, 2021.

\bibitem{Arcieri2023bridging}
G.\ Arcieri, C.\ Hoelzl, O.\ Schwery, D.\ Straub, K.G.\ Papakonstantinou, and
  E.\ Chatzi.
\newblock {Bridging POMDPs and Bayesian decision making for robust maintenance
  planning under model uncertainty: An application to railway systems}.
\newblock {\em Reliability Engineering and System Safety}, 239:109496, 2023.

\bibitem{Bull2021}
L.A.\ Bull, P.\ Gardner, J.\ Gosliga, T.J.\ Rogers, N.\ Dervilis, E.J.\ Cross,
  E.\ Papatheou, A.E.\ Maguire, C.\ Campos, and K.\ Worden.
\newblock {Foundations of population-based SHM, Part I: Homogeneous populations
  and forms}.
\newblock {\em Mechanical Systems and Signal Processing}, 148:107141, 2021.

\bibitem{Gosliga2021}
J.\ Gosliga, P.A.\ Gardner, L.A.\ Bull, N.\ Dervilis, and K.\ Worden.
\newblock {Foundations of Population-based SHM, Part II: Heterogeneous
  populations – Graphs, networks, and communities}.
\newblock {\em Mechanical Systems and Signal Processing}, 148:107144, 2021.

\bibitem{Gardner2021b}
P.\ Gardner, L.A.\ Bull, J.\ Gosliga, N.\ Dervilis, and K.\ Worden.
\newblock {Foundations of population-based SHM, Part III: Heterogeneous
  populations – Mapping and transfer}.
\newblock {\em Mechanical Systems and Signal Processing}, 148:107142, 2021.

\bibitem{Tsialiamanis2021}
G.\ Tsialiamanis, C.\ Mylonas, E.\ Chatzi, N.\ Dervilis, D.J.\ Wagg, and K.\
  Worden.
\newblock {Foundations of population-based SHM, Part IV: The geometry of spaces
  of structures and their feature spaces}.
\newblock {\em Mechanical Systems and Signal Processing}, 157:107692, 2021.

\bibitem{Poole2023negative}
J.\ Poole, P.\ Gardner, N~Dervilis, J.H.\ Mclean, T.J.\ Rogers, and K.\ Worden.
\newblock On negative transfer for transfer learning in dynamics.
\newblock {\em Proceedings of the 41st International Conference on Modal
  Analysis (IMAC-XLI)}, 2023.

\bibitem{Brennan2023comparison}
D.S.\ Brennan, E.J.\ Cross, and K.\ Worden.
\newblock A comparison of structural similarity metrics within population-based
  structural health monitoring.
\newblock {\em Structural Health Monitoring 2023}, 2023.

\bibitem{Poole2022alignment}
J.\ Poole, P.\ Gardner, N.\ Dervilis, L.\ Bull, and K.\ Worden.
\newblock On statistic alignment for domain adaptation in structural health
  monitoring.
\newblock {\em Structural Health Monitoring}, page 14759217221110441, 2022.

\bibitem{Gardner2022aircraft}
P.\ Gardner, L.A.\ Bull, J.\ Gosliga, J.\ Poole, N.\ Dervilis, and K.\ Worden.
\newblock {A population-based SHM methodology for heterogeneous structures:
  Transferring damage localisation knowledge between different aircraft wings}.
\newblock {\em Mechanical Systems and Signal Processing}, 172:108918, 2022.

\bibitem{Wang2019negative}
Z.\ Wang, Z.\ Dai, B.\ P{\'o}czos, and J.\ Carbonell.
\newblock Characterizing and avoiding negative transfer.
\newblock In {\em Proceedings of the IEEE/CVF Conference on Computer Vision and
  Pattern Recognition}, pages 11293--11302, 2019.

\bibitem{Worden2020}
K.\ Worden, L.A.\ Bull, P.\ Gardner, J.\ Gosliga, T.J.\ Rogers, E.J.\ Cross,
  E.\ Papatheou, W.\ Lin, and N.\ Dervilis.
\newblock {A brief introduction to recent developments in population-based
  structural health monitoring}.
\newblock {\em Frontiers in Built Environment}, 6:146, 2020.

\bibitem{Gardner2018adaptation}
P.\ Gardner, X.\ Lui, and K.\ Worden.
\newblock On the application of domain adaptation in structural health
  monitoring.
\newblock {\em Mechanical Systems and Signal Processing}, 138:106550, 2018.

\bibitem{Yu2020adaptation}
K.\ Yu, Q.\ Fu, H.\ Ma, T.R.\ Lin, and X.\ Li.
\newblock Simulation data driven weakly supervised adversarial domain
  adaptation approach for intelligent cross-machine fault diagnosis.
\newblock {\em Structural Health Monitoring}, 20(4):{2182--2198}, 2020.

\bibitem{Soleimani2023zeroshot}
M.H.\ Soleimani-Babakamali, R.\ Soleimani-Babakamali, K.\ Nasrollahzadeh, O.\
  Avci, S.\ Kiranyaz, and E.\ Taciroglu.
\newblock Zero-shot transfer learning for structural health monitoring using
  generative adversarial networks and spectral mapping.
\newblock {\em Mechanical Systems and Signal Processing}, 198:110404, 2023.

\bibitem{Cao2018preprocessing}
P.\ Cao, S.\ Zhang, and J.\ Tang.
\newblock Preprocessing-free gear fault diagnosis using small datasets with
  deep convolutional neural network-based transfer learning.
\newblock {\em IEEE Access}, 6:26241--26253, 2018.

\bibitem{Bull2023hierarchical}
L.A.\ Bull, D.\ Di~Francesco, M.\ Dhada, O.\ Steinert, T.\ Lindgren, A.K.\
  Parlikad, A.B.\ Duncan, and M.\ Girolami.
\newblock {Hierarchical Bayesian modeling for knowledge transfer across
  engineering fleets via multitask learning}.
\newblock {\em Computer-Aided Civil and Infrastructure Engineering},
  38(7):821--848, 2023.

\bibitem{Dardeno2023hierarchical}
T.A.\ Dardeno, R.S.\ Mills, N.\ Dervilis, K.\ Worden, and L.A.\ Bull.
\newblock On the hierarchical bayesian modelling of frequency response
  functions.
\newblock {\em arXiv preprint arXiv:2307.06263}, 2023.

\bibitem{Hughes2023decision}
A.J.\ Hughes, J.\ Poole, N.\ Dervilis, P.\ Gardner, and K.\ Worden.
\newblock {A decision framework for selecting information-transfer strategies
  in population-based SHM}.
\newblock {\em arXiv preprint arXiv:2307.06978}, 2023.

\bibitem{Kamariotis2022}
A.\ Kamariotis, E.\ Chatzi, and D.\ Straub.
\newblock {Value of information from vibration-based structural health
  monitoring extracted via Bayesian model updating}.
\newblock {\em Mechanical Systems and Signal Processing}, 166:108465, 2022.

\bibitem{Hughes2022}
A.J.\ Hughes, L.A.\ Bull, P.\ Gardner, R.J.\ Barthorpe, N.\ Dervilis, and K.\
  Worden.
\newblock {On risk-based active learning for structural health monitoring}.
\newblock {\em Mechanical Systems and Signal Processing}, 167:108569, 2022.

\bibitem{Hughes2022c}
A.J.\ Hughes, L.A.\ Bull, P.\ Gardner, N.\ Dervilis, and K.\ Worden.
\newblock {On robust risk-based active-learning algorithms for enhanced
  decision support}.
\newblock {\em Mechanical Systems and Signal Processing}, 181:109502, 2022.

\bibitem{Zhang2023quantifying}
W.H.\ Zhang, J.\ Qin, D.G.\ Lu, M.\ Liu, and M.H.\ Faber.
\newblock {Quantifying the value of structural health monitoring information
  with measurement bias impacts in the framework of dynamic Bayesian Network}.
\newblock {\em Mechanical Systems and Signal Processing}, 187:109916, 2023.

\bibitem{Brennan2021irreducible}
D.S.\ Brennan, J.\ Gosliga, E.J.\ Cross, and K.\ Worden.
\newblock On implementing an irreducible element model schema for
  population-based structural health monitoring.
\newblock {\em Structural Health Monitoring 2021}, 2021.

\bibitem{Kingma2014adam}
D.P.\ Kingma and J.\ Ba.
\newblock Adam: A method for stochastic optimization.
\newblock {\em arXiv preprint arXiv:1412.6980}, 2014.

\bibitem{Brennan2022quantifying}
D.S.\ Brennan, T.J.\ Rogers, E.J.\ Cross, and K.\ Worden.
\newblock On quantifying the similarity of structures via a graph neural
  network for population-based structural health monitoring.
\newblock {\em Proceedings of ISMA 2022-International Conference on Noise and
  Vibration Engineering and USD 2022-International Conference on Uncertainty in
  Structural Dynamics}, 2022.

\end{thebibliography}

\end{document}